\documentclass[runningheads]{llncs}

\usepackage{graphicx}
\usepackage{comment}
\usepackage{amsmath,amssymb}
\usepackage{color}
\usepackage{url}
\usepackage{hyperref}

\newif\ifreview
\reviewfalse

\ifreview
	\usepackage{lineno}

	\linenumbers
\fi

\usepackage[dvipsnames]{xcolor}

\usepackage{blindtext}
\usepackage{caption}
\usepackage{subcaption}
\usepackage{cleveref}
\crefname{section}{Sec.}{Secs.}
\Crefname{section}{Section}{Sections}
\Crefname{table}{Table}{Tables}
\crefname{table}{Tab.}{Tabs.}
\crefname{figure}{Fig.}{Figs.}
\Crefname{figure}{Figure}{Figures}

\usepackage{booktabs}
\usepackage{multicol}
\usepackage{makecell}

\makeatletter
\newcommand{\printfnsymbol}[1]{%
  \textsuperscript{\@fnsymbol{#1}}%
}
\makeatother

\begin{document}

\def\SubNumber{8}

\def\GCPRTrack{Main Track}

\title{The Impact of Synthetic Data on Object Detection Model Performance: A Comparative Analysis with Real-World Data}

\ifreview
\else
	\titlerunning{The Impact of Synthetic Data on Object Detection Model Performance}

	\author{Muammer Bay\inst{1,2}\and
	Timo von Marcard\inst{2}\and
    Dren Fazlija\inst{3}}
	
	\authorrunning{M. Bay et al.}

        \institute{LycheeAI, Hannover, Germany	\and Hochschule Hannover, Germany \and L3S Research Center, Leibniz University Hannover, Germany}
    
\fi

\maketitle              %

\begin{abstract}
Recent advances in generative AI, particularly in computer vision (CV), offer new opportunities to optimize workflows across industries, including logistics and manufacturing. However, many AI applications are limited by a lack of expertise and resources, which forces a reliance on general-purpose models. Success with these models often requires domain-specific data for fine-tuning, which can be costly and inefficient.  Thus, using \textit{synthetic data} for fine-tuning is a popular, cost-effective alternative to gathering real-world data. This work investigates the impact of synthetic data on the performance of object detection models, compared to models trained on real-world data only, specifically within the domain of warehouse logistics. 
To this end, we examined the impact of synthetic data generated using the NVIDIA Omniverse Replicator tool on the effectiveness of object detection models in real-world scenarios. It comprises experiments focused on pallet detection in a warehouse setting, utilizing both real and various synthetic dataset generation strategies.  Our findings provide valuable insights into the practical applications of synthetic image data in computer vision, suggesting that a balanced integration of synthetic and real data can lead to robust and efficient object detection models.

\end{abstract}

\section{Introduction}

The demand for extensive datasets to train sophisticated models is ever-increasing in the rapidly evolving fields of artificial intelligence (AI) and computer vision (CV). 
However, collecting real-world data is costly, time-consuming, and fraught with potential biases and privacy concerns. Furthermore, capturing the diversity of scenarios needed for comprehensive model training is often unfeasible in real-world settings. 
Synthetic data generated through platforms such as NVIDIA’s Omniverse Replicator~\cite{Nvidia2025replicator} provides a promising solution for creating varied and complex AI training scenarios. 
This form of data potentially sidesteps some logistical and ethical issues associated with real-world data collection, offering a bias-reduced and privacy-compliant training alternative, which may enhance the adaptability and generalizability of AI models. 
As such, Gartner estimates that by 2030, synthetic data will surpass real data in AI models~\cite{Gartner2022}, highlighting its rising significance.
In warehouse logistics, where AI advancements have been notable since the 2000s~\cite{di2023machine,naumann2023literature,Yang2021WarehouseMM}, small and medium-sized enterprises (SMEs) often lack the resources to leverage AI technologies fully.
Hence, we believe that SMEs – particularly start-ups – would greatly benefit from synthetic data.
To assess the effectiveness and utility of synthetic data for low-resource applications of computer vision models, we examined a specific logistics use case: detecting pallets in a warehouse environment.
Pallet detection is crucial, as accuracy and efficiency are vital for computer vision tasks such as automated load handling, which can enhance warehouse operations.
This use case also provides a realistic setting to rigorously test the capabilities of existing object detection models while addressing common challenges, such as dealing with out-of-distribution data, i.e., data that is whose statistical properties do not align with our training data.
In this context, such distributional shifts may represent changes in the environment (e.g., can we adapt an existing system for use in a different warehouse or even outside a warehouse setting?) or variations in objects (such as differences in color, material, or pallet size).

\begin{figure*}[h]
    \centering
    \includegraphics[width=0.75\linewidth]{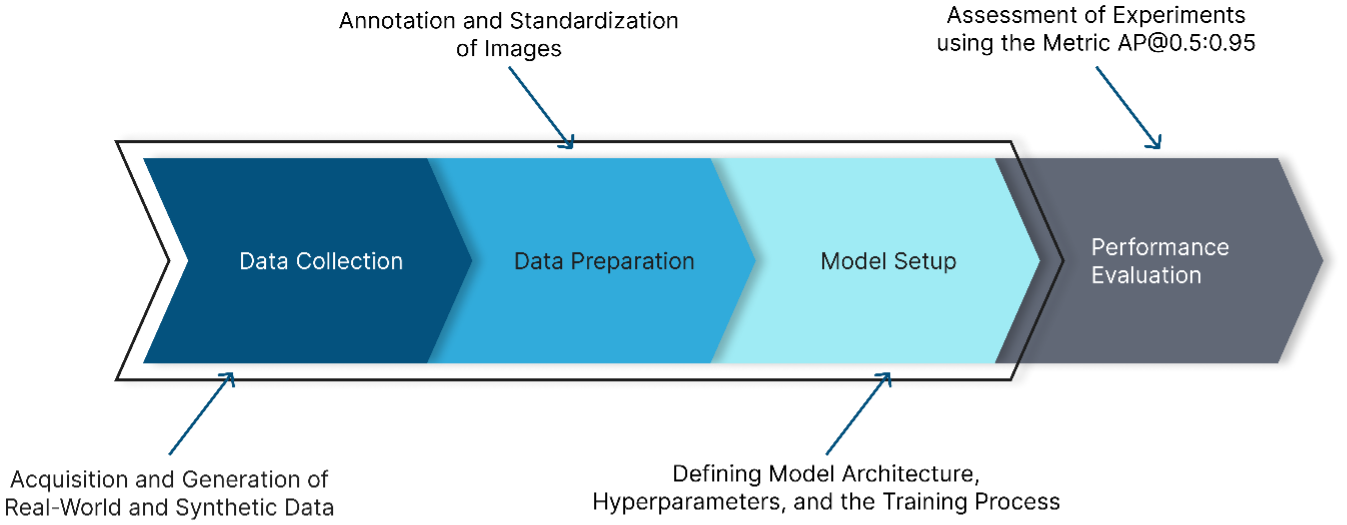}
    \caption{Overview of our Four-Stage Experimental Workflow. The process begins with acquiring real-world and synthetic data, followed by data preparation and processing. In the model setup phase, models are configured and trained. Afterwards, we assess the fine-tuned models against various benchmarks.}
    \label{fig:workflow}
\end{figure*}

\paragraph{Contributions.}
In this work, we examine the impact of synthetic data on the accuracy, robustness, and generalization of an object detection model, focusing on varying levels of data realism and fine-tuning techniques. 
Following the experimental workflow visualized in~\cref{fig:workflow}, we aim to: $(i)$ determine whether synthetic data can serve as a partial substitute for real-world environments, $(ii)$ analyze how the realism of synthetic images influences the model's performance on both in-distribution and out-of-distribution data, and $(iii)$ identify the most effective fine-tuning procedures for incorporating synthetic data. 
Our findings highlight intriguing nuances in model performance, revealing the benefits and drawbacks of varying data realism and fine-tuning strategies.

\paragraph{Reproducibility.} We provide training configs, Omniverse Replicator scripts, the full asset manifest, random seeds, and trained checkpoints at GitHub\footnote{\url{https://github.com/MuammerBay/omniverse-replicator-sim2real-analysis}} and Zenodo\footnote{\url{https://doi.org/10.5281/zenodo.17308406}}. Environment details (OS, CUDA, PyTorch/torchvision) and exact hyperparameters are in the repository.

\section{Related Work}

\subsection{Synthetic Data Generation via 3D Engines}
With the rise of powerful image generation models, particularly GANs~\cite{goodfellow2014generative} and Diffusion Models~\cite{sohl-dickstein15,ho2020denoising}, employing synthetic image data for developing machine learning models has become an increasingly accessible option. 
However, generating photorealistic snapshots of isolated items is insufficient for emulating object classification tasks that unfold in structured settings such as warehouses. 
Scene context, object-object interactions, and lighting dynamics all influence the appearance and recognisability of a target object as strongly as its geometry or texture.
2D generators struggle to capture these factors because they lack an explicit representation of three‑dimensional space.
Adjacent research, therefore, pivots to physics‑based 3D engines to simulate real-world, hyper-realistic scenarios.
These engines are also accompanied by additional tools that allow researchers to generate relevant computer vision metadata, such as bounding boxes or segmentation masks.

Unity~\cite{unity2025}, a game engine for creating immersive 3D environments, is a prominent tool for synthesizing highly realistic scenes for computer vision tasks.
Practical applications of Unity involve semantic segmentation of urban scenes to train computer vision models for autonomous driving~\cite{ros2016synthia}, as well as generating data for human-centric computer vision models that detect people and analyze human poses~\cite{ebadi2022psphdri}.
Many adjacent publications build upon the NVIDIA platforms Deep Learning Data Synthesizer~\cite{to2018ndds} and Omniverse Replicator~\cite{Nvidia2025replicator}. 
While researchers have successfully used the former tool to generate synthetic data (e.g.,~\cite{jalal2019sidod}), most of the more novel work builds upon NVIDIA's Omniverse Replicator. 
It produces high-fidelity datasets for AI applications, providing metadata like bounding boxes. 
Additionally, its integration with Isaac Sim makes Omniverse Replicator a pivotal resource in logistics and industrial applications, as evidenced by recent work on digital twins~\cite{nassif2024creating,zhang2022automatic} and medical decision support applications~\cite{hydock2023generation}.
Considering its continuous development, growing dominance, and impact in these fields, we used the Omniverse Replicator to generate 3D synthetic warehouse scenes for our pallet detection task.

\subsection{Synthetic vs. Real Data}

\textbf{The Domain Gap.}
One core issue of utilizing synthetic data is the domain gap (also called the simulation-to-reality gap), which refers to the differences in characteristics between synthetic and real-world data.
Variations in properties such as lighting, textures, object placements, and overall scene composition cause these differences, which can hinder the performance of models trained on synthetic data when applied to real-world scenarios.
To address this challenge, several studies have found that domain randomization (i.e., introducing a variety of random variations into synthetic data) enhances the model's ability to generalize from synthetic to real data, effectively closing the domain gap \cite{Vanherle2022BMVC,knitt2022palletpose}.
This is especially relevant in out-of-distribution (OOD) scenarios, where the model encounters real-world conditions that were not seen during training. In this work, we evaluate both in-distribution and OOD performance to assess the generalization power of synthetic training data.

\textbf{Strategies for Effective Model Training.}
Previous studies have investigated various training approaches, including mixed training, which combines synthetic and real data to enhance model performance. 
However, one strategy that has gained prominence is transfer learning, a cornerstone of modern AI development.
Transfer learning aims to improve the performance of models on specific tasks by leveraging knowledge acquired from related yet distinct domains.
While one may be inclined to simply simultaneously train on synthetic and real training (referred to as "mixing"), Li et al.~\cite{li2024synthetic} introduced a two-stage fine-tuning framework called "bridged transfer," where models are first trained on synthetic data and then fine-tuned with real data. 
This approach demonstrates higher accuracy than other methods, highlighting the benefits of leveraging synthetic data in the initial training phase.
Accordingly, we compare two integration schemes: the naïve mixing method and bridged transfer learning (BTL).
Mirroring the protocol of~\cite{ebadi2022psphdri}, we also investigate a range of synthetic‑to‑real ratios to pinpoint the blend that yields the highest downstream accuracy, allowing us to reason about different scenarios (e.g., having restricted access to real-world data).
Recent work also explores diffusion-based data generation and compositing for detection (e.g.,~\cite{feng2024instagen}), as well as modern detectors (DETR variants; open-set pretraining such as Grounding DINO~\cite{liu2024grounding}). 
We scope this study to Faster R-CNN to control confounds and match SME deployment constraints (mature toolchain, permissive license), leaving broader model families and generative pipelines for future work.

\section{Dataset Creation Process}

\subsection{Real-World Data Acquisition}\label{sec:realworlddata}
\begin{figure*}
    \centering
    \begin{subfigure}{0.45\linewidth}
        \centering
        \includegraphics[width=0.75\linewidth]{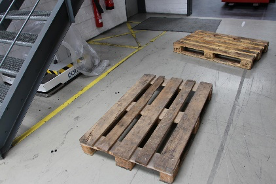}
    \end{subfigure}
    \hspace{0.05cm}
    \begin{subfigure}{0.45\linewidth}
        \centering
        \includegraphics[width=0.75\linewidth]{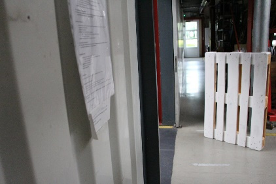}
    \end{subfigure}
    \caption{A comparison between an in-distribution sample (left) and an out-of-distribution image (right). The latter image displays a pallet in an unusual color and orientation.}
    \label{fig:idvsood}
\end{figure*}

Most SMEs struggle with limited access to relevant image data. To address this, we simulated a low-resource environment for pallet detection by capturing 220 photos in a local industrial warehouse. Of these, 200 images serve as our baseline, divided into training (160), test (20), and validation (20) datasets, depicting consistent pallets. The remaining 20 images represent out-of-distribution scenarios with different colors, orientations, and increased occlusion. All images (summarized in~\cref{tab:realdatasplits}) were taken at a high resolution (5184x3456 pixels) with an 18mm focal length. 
\Cref{fig:idvsood} shows an example of a test image and an out-of-distribution data point.

\begin{table}[]
    \centering
    \caption{Overview of the Different Real-World Data Splits.}
    \label{tab:realdatasplits}
    \begin{tabular}{lcl}
        \toprule
        Split & No. Images & \makecell[c]{Notes} \\\midrule
        Train (ID) & 160 & Real-World Training Data for Fine-tuning\\
        Val (ID) & 20 & Validation Set, Shares Core Characteristics with Train\\
        Test (ID) & 20 & Unseen Data, Same Characteristics as Train and Val\\
        Test (OOD) & 20 & Different colors/orientations \& increased occlusion\\\bottomrule
    \end{tabular}
\end{table}

\subsection{Synthetic Data Generation}
Our data generation pipeline utilizes NVIDIA's Omniverse Replicator~\cite{Nvidia2025replicator}, alongside the Omniverse Nucleus Server~\cite{Nvidia2025Nucleus}, to create three synthetic datasets: "Realistic," "Half-Realistic," and "Random." 
Each dataset progressively randomizes key properties of pallets and environmental factors, allowing us to analyze distribution shifts in training signals. 
All datasets consist of 500 rendered images at a resolution of 1536x1024 pixels, with a normally distributed focal length (mean: 18mm, standard deviation: 4mm), and employ ray tracing for 3D environments.
We refer to the "Replicator" directory of our codebase for implementation details.

\begin{figure*}
    \centering
    \begin{subfigure}{0.45\linewidth}
        \centering
        \includegraphics[width=0.75\linewidth]{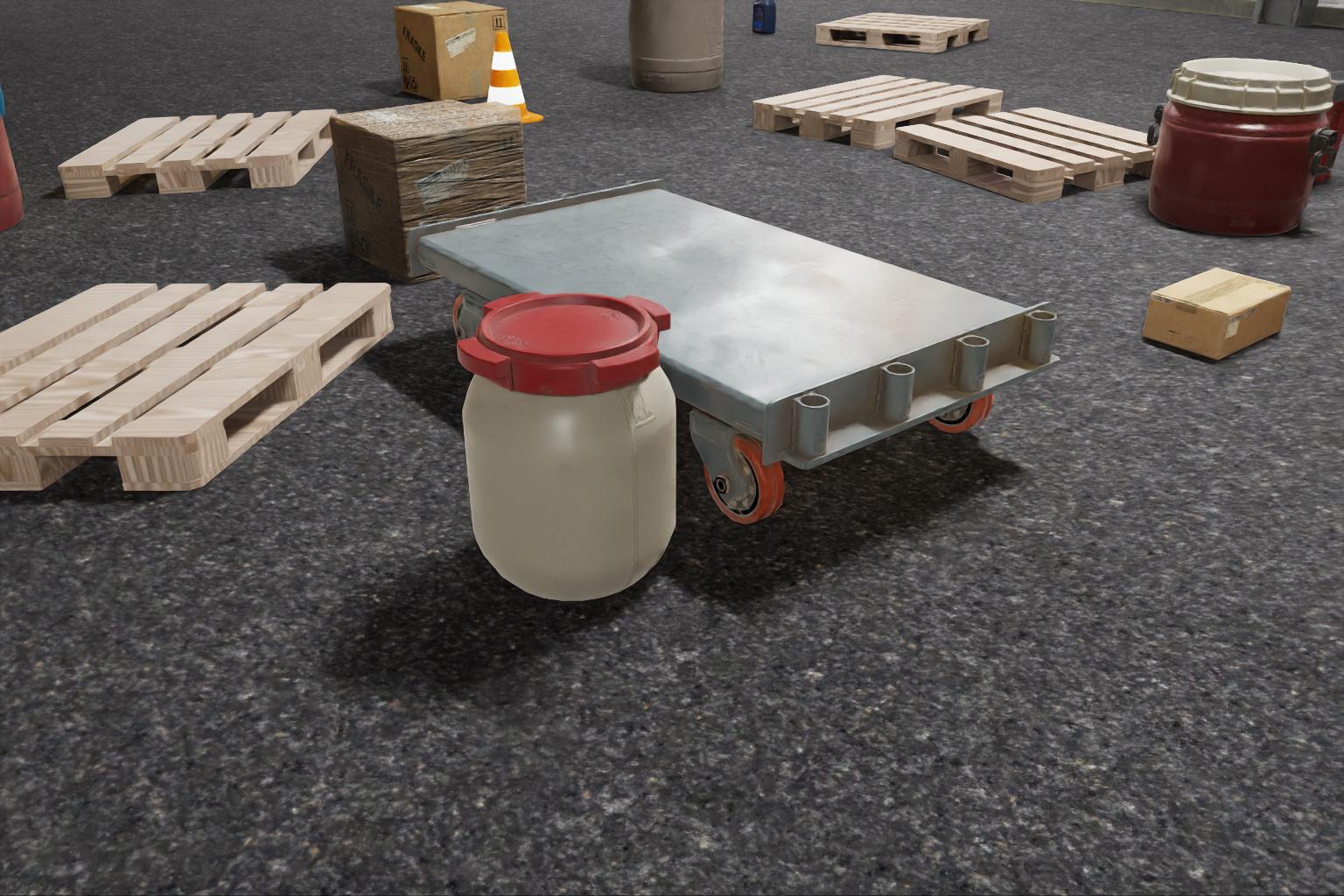}
    \end{subfigure}
    \hspace{0.05cm}
    \begin{subfigure}{0.45\linewidth}
        \centering
        \includegraphics[width=0.75\linewidth]{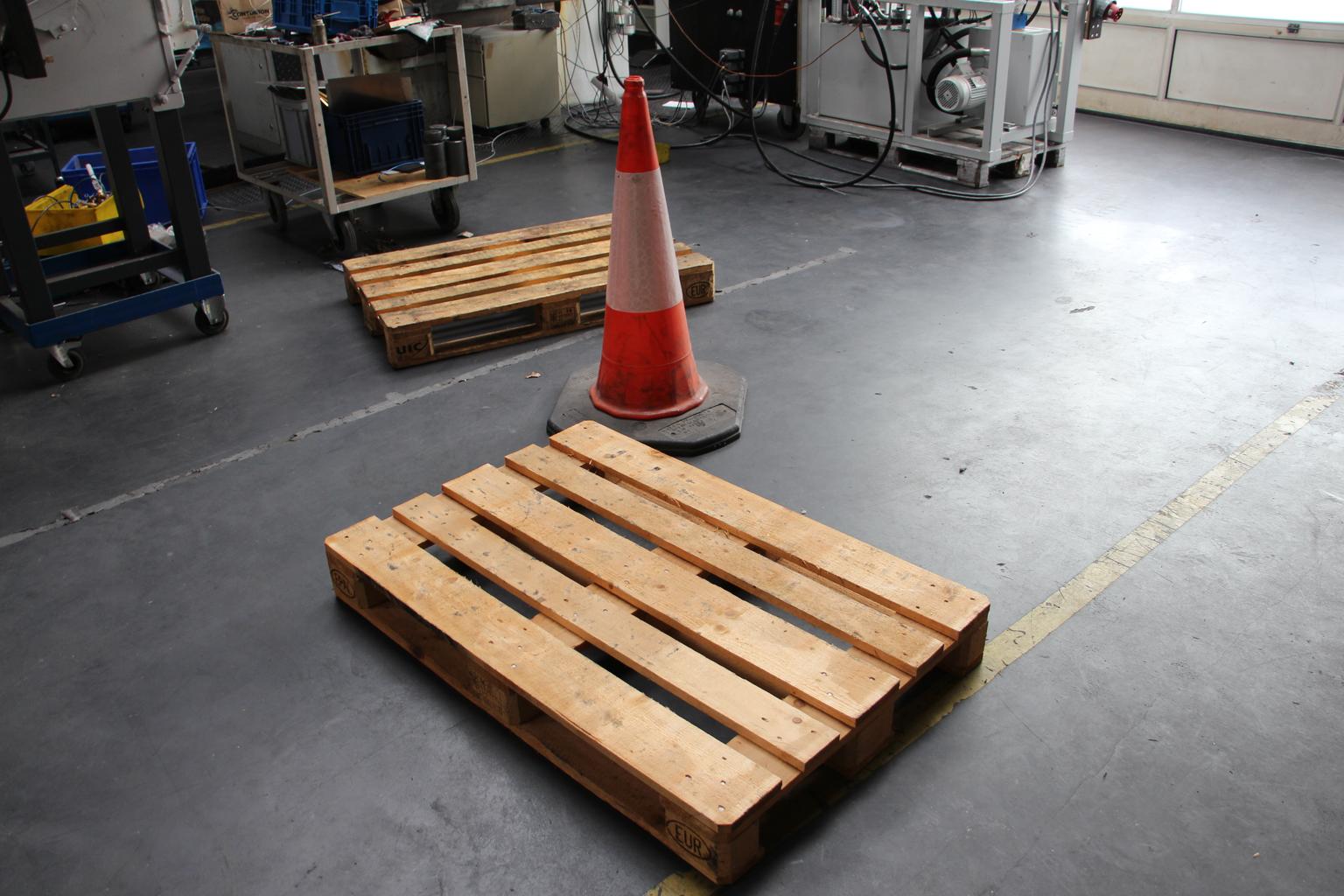}
    \end{subfigure}
    \caption{A comparison between a realistic synthetic image (left) and a real-world image (right). }
    \label{fig:realisticvsreal}
\end{figure*}

\textbf{Realistic Synthetic Data.} 
We created a realistic dataset that mimics real-world conditions with accurate textures, lighting, and object placements. 
The scenes feature appropriately sized pallets in a logistics warehouse, along with distractor objects like cardboard boxes and pushcarts.
Lighting is designed to produce realistic shadows and variations, simulating different times of day.
A camera captures the scene from randomized viewpoints while textures, colors, and material properties (e.g., roughness) vary and objects are scattered, producing a unique layout in each frame.
\Cref{fig:realisticvsreal} shows the comparison of the realistic synthetic image and the real-world image of a pallet.

\begin{figure*}
    \centering
    \begin{subfigure}{0.45\linewidth}
        \centering
        \includegraphics[width=0.75\linewidth]{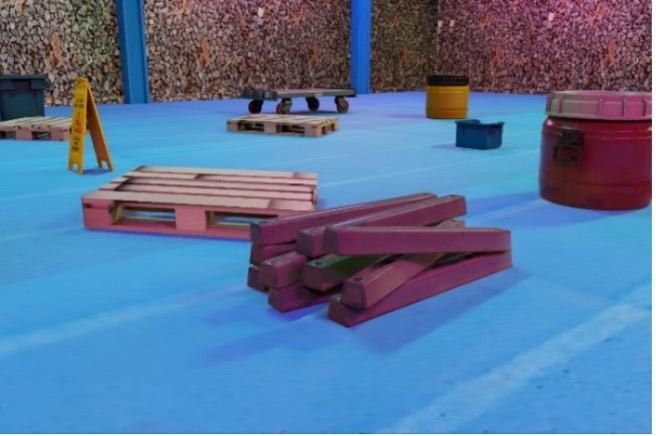}
    \end{subfigure}
    \hspace{0.05cm}
    \begin{subfigure}{0.45\linewidth}
        \centering
        \includegraphics[width=0.75\linewidth]{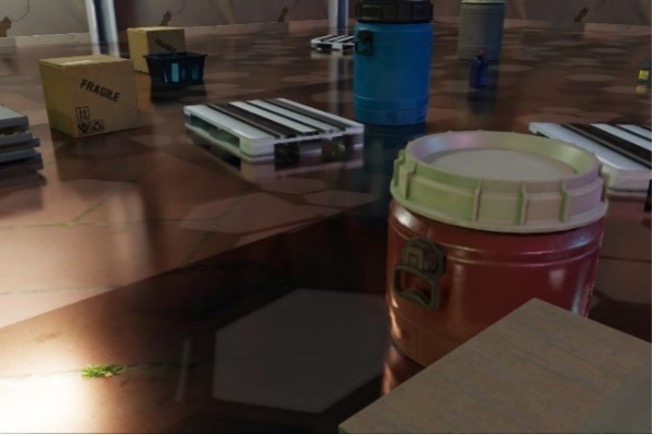}
    \end{subfigure}
    \caption{Examples of Half-Realistic Synthetic Images.}
    \label{fig:halfrealisticexamples}
\end{figure*}

\noindent\textbf{Half-Realistic Synthetic Data.}
The Half-Realistic dataset aims to balance realism and variability by incorporating partially stylized elements while keeping some realistic features. 
This approach is designed to help models generalize better by exposing them to variations not found in real-world data.
It includes a range of realistic and randomized textures, pallets, warehouse distractor objects, and diverse floor and wall textures.
Lighting conditions vary in color and intensity, with some lights randomly turned off. 
Additionally, objects and textures are randomized to create partially abstract scenes, mixing realistic and stylized elements.
\Cref{fig:halfrealisticexamples} shows examples of Half-Realistic synthetic images.

\begin{figure*}
    \centering
    \begin{subfigure}{0.45\linewidth}
        \centering
        \includegraphics[width=0.75\linewidth]{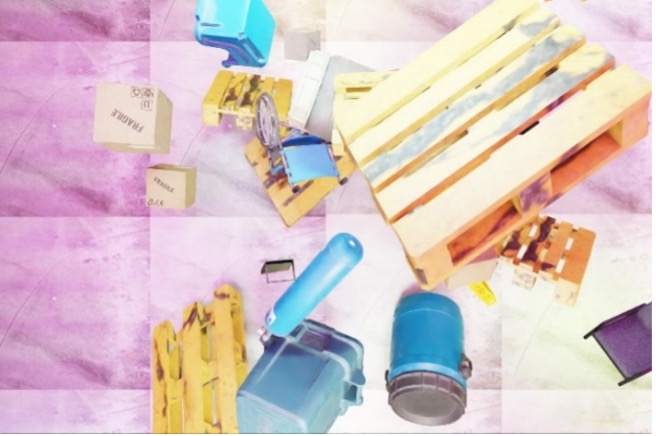}
    \end{subfigure}
    \hspace{0.05cm}
    \begin{subfigure}{0.45\linewidth}
        \centering
        \includegraphics[width=0.75\linewidth]{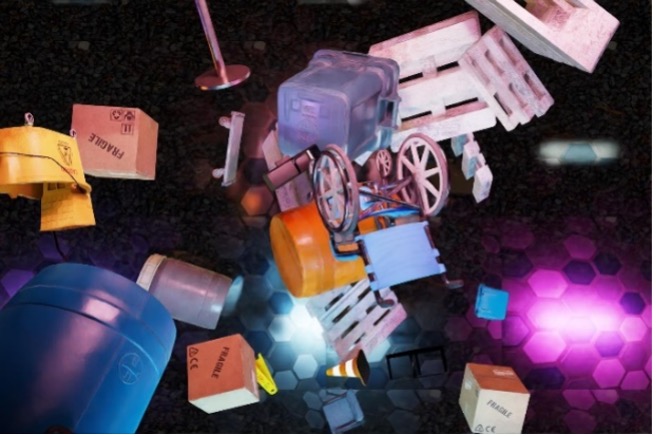}
    \end{subfigure}
    \caption{Examples of Random Synthetic Images.}
    \label{fig:randomexamples}
\end{figure*}

\noindent\textbf{Random Synthetic Data.}
This dataset introduces substantial variability and randomness to expose the model to a wide range of scenarios, including highly unlikely or rare events that occur in the real world.
We create diverse scenes by incorporating a variety of textures and unusual patterns, such as animal fur or artificial grass.
Additionally, we include non-typical items like wheelchairs and televisions, while also randomizing material properties such as roughness, metallic characteristics, and emissiveness.
To simulate extreme conditions, we heavily randomize lighting by adjusting the position, color, intensity, and visibility of lights.
Camera settings also undergo considerable variation, including changes in position, focal length, and focus distance.
\Cref{fig:randomexamples} shows examples of Random synthetic images.

\subsection{Data Processing}

\begin{figure}
    \centering
    \begin{subfigure}{0.45\linewidth}
        \centering
        \includegraphics[width=0.75\linewidth]{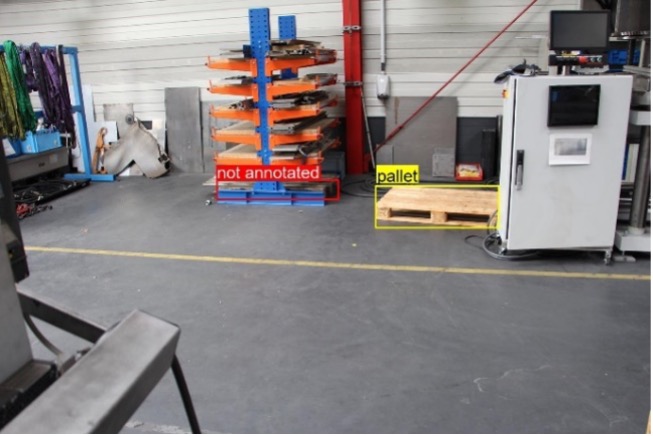}
    \end{subfigure}
    \hspace{0.05cm}
    \begin{subfigure}{0.45\linewidth}
        \centering
        \includegraphics[width=0.75\linewidth]{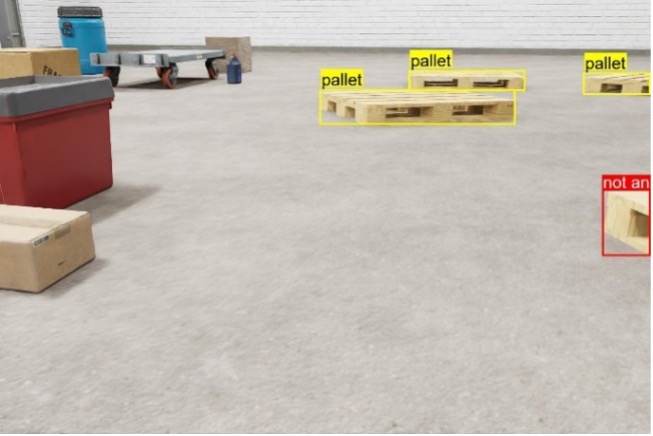}
    \end{subfigure}
    \caption{Comparison of real (left) and synthetic (right) data annotations. The annotations, including bounding boxes and semantic labels, are highlighted in yellow for objects meeting the 25\% visibility threshold, while those not meeting the threshold are shown in red, indicating no annotation.}
    \label{fig:annotationexamples}
\end{figure}

\textbf{Image Annotation.}
We used the open-source tool CVAT~\cite{CVAT2023} to create bounding boxes for pallets in our real-world data, defining each box by the pixel coordinates of the top-left and bottom-right corners. 
Boxes with less than 25\% visibility were omitted to ensure data quality.
Annotations were saved in text files, containing coordinates and labels for each image.
For synthetic data, we utilized Omniverse Replicator's annotation pipeline to automatically generate bounding boxes and semantic labels, reducing human error and ensuring accuracy. 
We also maintained consistent labeling conventions and a visibility threshold of 25\% for both data types.
\Cref{fig:annotationexamples} shows bounding boxes for both real and synthetic pallets and highlights pallets that fall below the threshold.

\noindent\textbf{Data Preparation.}
To maintain consistency in data handling during fine-tuning, we matched synthetic and real-world image files with their corresponding annotations, standardized the image size to 1536x1024 pixels, and ensured the image format was JPG. 
Furthermore, we processed the standardized images by performing normalization and further downscaling the images to 1200x800 pixels before forwarding the data to our model. 
Finally, we conducted manual quality checks to ensure the accuracy of annotations and eliminated any duplicate or corrupt data.

\section{Experiments}

\subsection{The Object Detection Model}
When selecting a baseline detector, we prioritized three requirements: strong out‑of‑the‑box accuracy, ease of deployment for SMEs, and a permissive license. 
The open‑source Python implementation of the Faster R‑CNN~\cite{ren2015faster}, which was trained on MS-COCO~\cite{lin2014microsoft}, satisfies all three. 
It is mature, well‑documented, and widely adopted, consistently performing well across different benchmarks~\cite{kim2020comparison,liu2020deep} while remaining free for commercial use. 
Therefore, we adopted Faster R‑CNN as the reference model for all subsequent experiments.

\textbf{Hyperparameters.}
We performed all subsequent training phases of our base model via Stochastic Gradient Descent~\cite{robbins1951stochastic} with a learning rate of 0.001, a momentum of 0.9, and an L2 regularization of 0.0005. 
We chose a batch size of 8 to accommodate the relatively small dataset sizes while effectively normalizing and processing each batch. 
We trained our baseline model for a maximum of 100 epochs and employed an early stopping mechanism, which ends training if performance does not improve for 15 consecutive epochs.
For our analysis, we performed the following experiments across 10 seeds.

\textbf{Compute \& Generation budget.} Synthetic generation was run on a personal PC with a single RTX 4090, where producing 500 images took ca. 2 hours (ca. 250 images/hour). 
Fine-tuning per seed was run on a single H100 and took ca. 19 hours; across 10 seeds, this amounts to ca. 190 GPU-hours.

\subsection{Experimental Setup}

\begin{figure}
    \centering
    \includegraphics[width=0.65\linewidth]{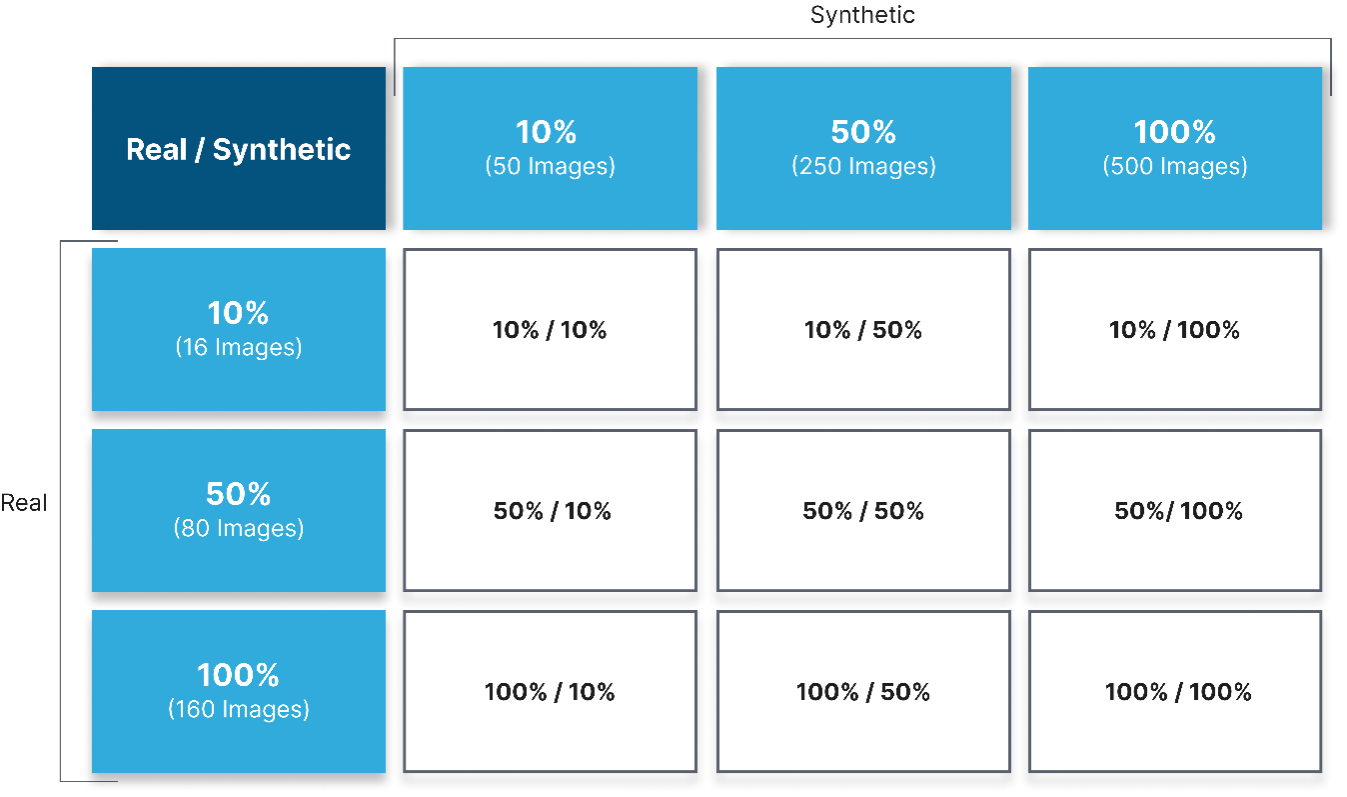}
    \caption{Data Utilization Ratios in Experimental Setup. This chart outlines the combination of real and synthetic data used in various proportions (10\%, 50\%, 100\%) for training object detection models.}
    \label{fig:ratios}
\end{figure}

To investigate the effects of synthetic data, we first fine-tuned our pre-trained Faster R-CNN model with our real-world training data (160 images).
We then compared its performance against different synthetic-data training regimens.
We also adjusted the ratios of our different datasets, specifically the relative amounts of data taken from each real-world and synthetic dataset. 
From this point forward, we will refer to these ratios to describe the composition of real-world and synthetic images within a training dataset (see~\cref{fig:ratios}). 
For example, a dataset with 10\% real and 10\% synthetic data contains 16 real images and 50 synthetic images.
For each data ratio, we used the same fixed subset of real and synthetic images across all runs. 
Only model initialization parameters were randomized, ensuring consistent dataset composition for fair performance comparison.

\noindent\textbf{Synthetic Data Only.}
We first trained Faster R-CNN solely with different ratios of Realistic, Half-Realistic, and Random synthetic data to examine whether synthetic data could substitute real-world data entirely.

\textbf{Mixed Method.}
To test whether we could \textit{reduce} the required real data substantially by introducing larger ratios of synthetic images to the fine-tuning dataset, we first assessed the mixed method, i.e., we \textit{simultaneously} train the model on both real and synthetic data. 
The hypothesis is that combining real and synthetic data throughout training can improve the model's generalization ability across different scenarios.

\textbf{Bridged Transfer Learning.}
Recent studies, including the work by Li et al.~\cite{li2024synthetic}, suggest that using the mixed method may not be the most efficient approach for combining synthetic and real-world data. 
Instead, Li et al.~\cite{li2024synthetic} recommend the bridged transfer learning (BTL) strategy, which involves initially fine-tuning a model with synthetic images and then fine-tuning it with real images.
This two-stage process allows the model to learn general patterns and features from the synthetic data before refining its understanding with more complex and nuanced real-world data.

\textbf{Evaluation Metrics.}
Following the MS COCO benchmark~\cite{lin2014microsoft}, we will use the average precision (AP) over a 101-point interpolation of recall points $r \in \{0, 0.1, 0.2, \allowbreak \ldots, 1.0\}$ to describe the object detection performance for a specified Intersection over Union (IoU) threshold~\cite{jaccard1902lois}.
The IoU, also known as the Jaccard Index, allows us to quantitatively assesses how closely the predicted bounding box matches the actual object and defines correct detections when the IoU is above a specified threshold. 
To summarize the model's performance across multiple IoU thresholds, we will use the comprehensive metric AP@0.5:0.95, which averages AP across multiple IoU thresholds ranging from 0.5 to 0.95 in increments of 0.05, i.e.,
\begin{equation*}
    AP@0.5:0.95 = \frac{1}{10}\sum_{k=0}^{9}AP@(0.5 + 0.05 \cdot k)
\end{equation*}
This approach provides a detailed evaluation of a model's performance by considering different levels of localization accuracy, ensuring the model's ability to detect objects at varying degrees of strictness. 
We evaluate the models' performance on unseen in-distribution data on the real-world test dataset.
At the same time, we use the real-world OOD test dataset to assess the models' ability to generalize in out-of-distribution scenarios.
See~\cref{sec:realworlddata} for details about the two test datasets.
Additionally, for the baseline (cf.~\cref{tab:baseline}), we report two-sided 95\% confidence intervals (CIs) computed as
\[\bar{x}\ \pm\ t_{0.975,\,n-1}\,\frac{s}{\sqrt{n}},\]
where $\bar{x}$ is the sample mean, $s$ is the sample standard deviation ($ddof=1$), $n$ is the number of seeds (here $n=10$), and $t_{0.975,\,n-1}$ is the 97.5th percentile of Student's t-distribution with n-1 degrees of freedom~\cite{student1908probable,neyman1937outline}.

\section{Results}

In this section, we will continually reference the model's performance after fine-tuning with varying ratios of real-world data. Each figure in the subsequent subsections shows the baseline metrics from~\cref{tab:baseline}, represented by colored dashed lines: red for 10\% of real data, green for 50\%, and blue for 100\%.
All ratio values refer to static dataset sizes, where the real-world dataset contains 160 images and the synthetic dataset contains 500 images.
These baselines are fixed reference points for comparison and are not part of the actual experimental variation shown in the subsequent figures.
Furthermore, we will henceforth refer to the in-distribution testing set as "test data" and the out-of-distribution testing set as "OOD data".

\begin{table}[h!]
    \scriptsize
    \centering
    \renewcommand{\arraystretch}{1.2}
    \caption{AP@0.5:0.95 baseline performance of our Faster R-CNN when fine-tuned with only real-world images on our test and OOD data \textbf{across 10 seeds}. Values are reported as mean~$\pm$~SD and 95\% CI (over $n=10$).}
    \begin{tabular}{lcccccc}
        \toprule
        & \multicolumn{6}{c}{\textbf{Real-World Ratios}} \\
        \cmidrule(lr){2-7}
        & \multicolumn{2}{c}{\textbf{10\%}} & \multicolumn{2}{c}{\textbf{50\%}} & \multicolumn{2}{c}{\textbf{100\%}} \\
        \cmidrule(lr){2-3}\cmidrule(lr){4-5}\cmidrule(lr){6-7}
        \textbf{Test Set} & \textbf{AP $\pm$ SD} & \textbf{95\% CI} & \textbf{AP $\pm$ SD} & \textbf{95\% CI} & \textbf{AP $\pm$ SD} & \textbf{95\% CI} \\
        \midrule
        Test Data (ID) &
        $0.643 \pm 0.024$ & [0.627, 0.659] &
        $0.789 \pm 0.016$ & [0.779, 0.800] &
        $0.795 \pm 0.010$ & [0.788, 0.801] \\
        ODD Data &
        $0.466 \pm 0.041$ & [0.438, 0.494] &
        $0.581 \pm 0.013$ & [0.572, 0.590] &
        $0.585 \pm 0.024$ & [0.569, 0.600] \\
        \bottomrule
    \end{tabular}
    \label{tab:baseline}
\end{table}

\subsection{Synthetic Data Only}
\begin{figure}[h]
    \centering
    \begin{subfigure}[b]{0.49\linewidth}
        \centering
        \includegraphics[width=1.0\linewidth]{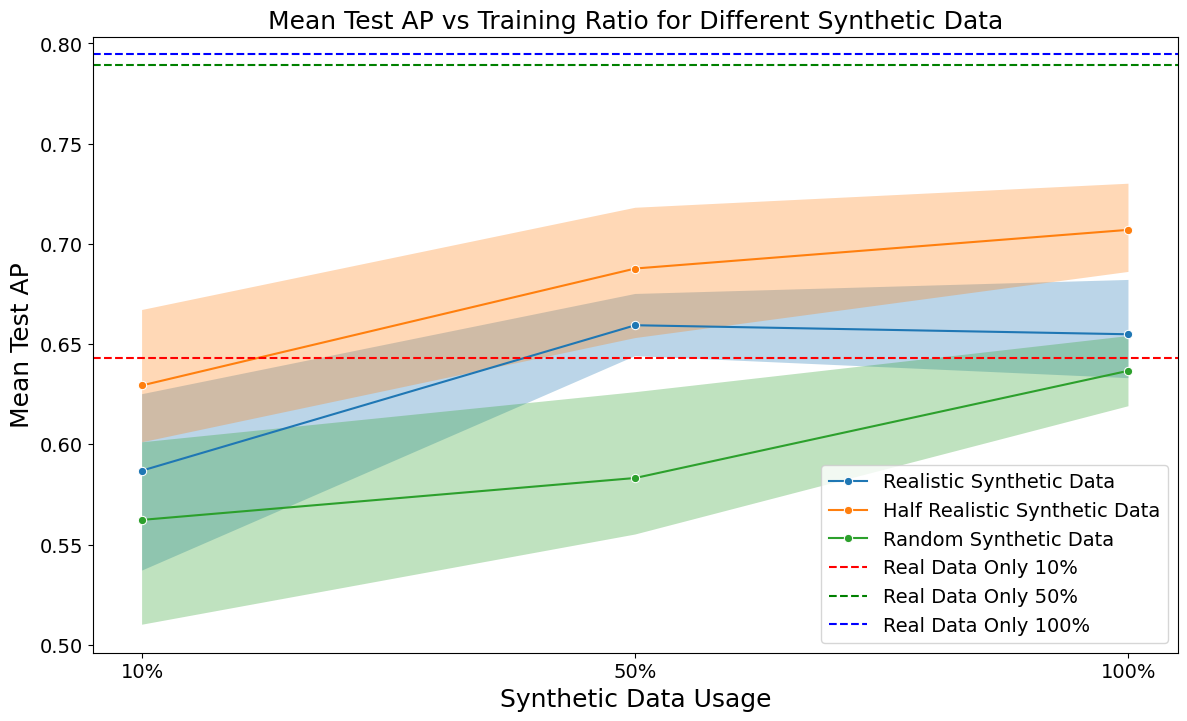}
        \caption{Test Performance}
        \label{fig:syntheticonlytest}
    \end{subfigure}
    \hfill
    \begin{subfigure}[b]{0.49\linewidth}
        \centering
        \includegraphics[width=1.0\linewidth]{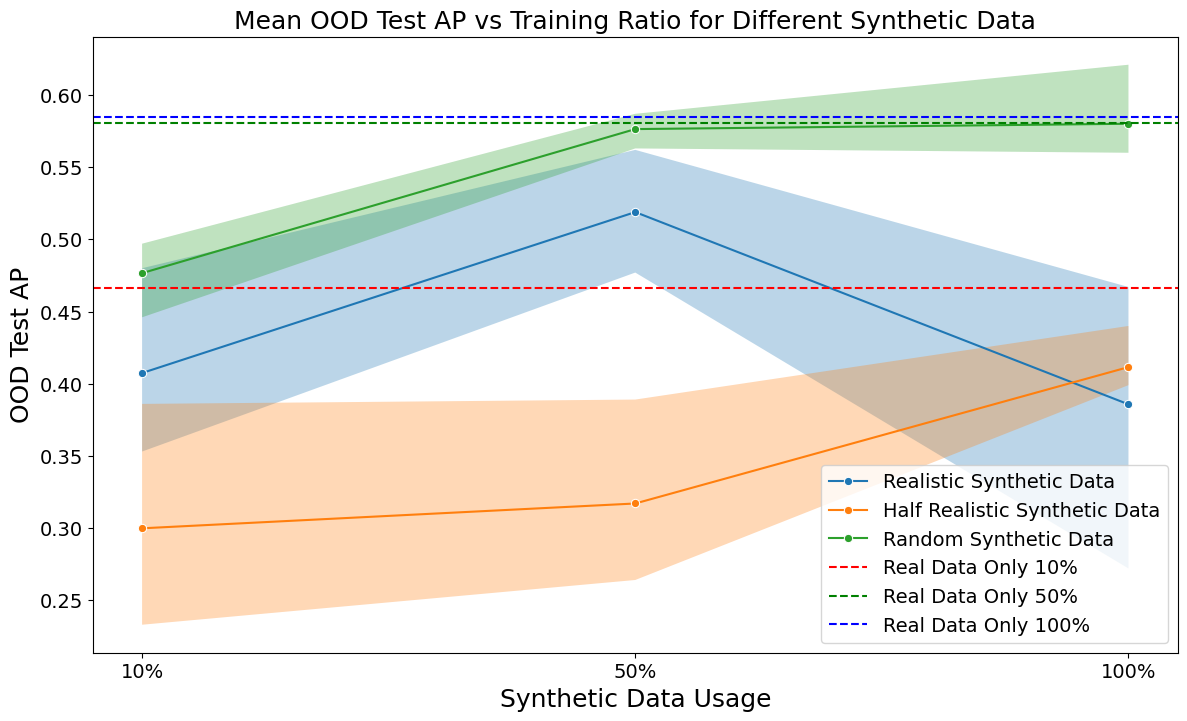}
        \caption{OOD Data Performance}
        \label{fig:syntheticonlyood}
    \end{subfigure}
    \caption{Faster R-CNN performance on the test and OOD dataset after fine-tuning with different amounts of only synthetic data.}
    \label{fig:syntheticonly}
\end{figure}

Despite having access to more data points when training on synthetic data (up to 500 synthetic images versus 160 real images), \textit{exclusively} training or fine-tuning on synthetic data is insufficient for our real-world application.
As visualized in~\cref{fig:syntheticonlytest}, Faster R-CNN models consistently perform far worse than their real-data-only counterparts on in-distribution test data, at best outperforming the 10\% real-data-only instance of the model.
The OOD test performance of synthetic-only models (\cref{fig:syntheticonlyood}) shows that, while they still perform worse overall than their real-data-only counterparts, some Faster R-CNN instances benefit from fine-tuning on synthetic data only.
Training exclusively on randomly generated pallet images yields the strongest OOD performance among the synthetic-only settings.

\subsection{Mixed Data Approach}

\begin{figure}[h]
    \centering
    
    \begin{subfigure}[b]{0.49\linewidth}
        \centering
        \includegraphics[width=1.0\linewidth]{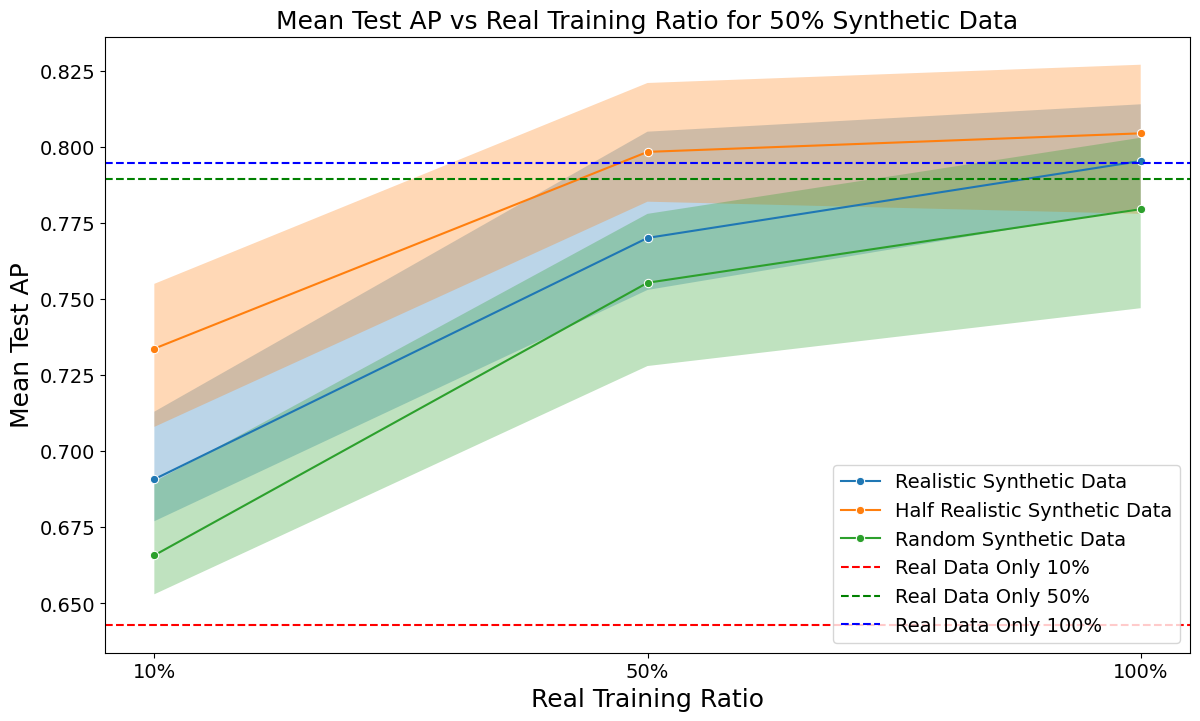}
        \caption{Test Performance}
        \label{fig:mixedtest}
    \end{subfigure}
    \hfill
    \begin{subfigure}[b]{0.49\linewidth}
        \centering
        \includegraphics[width=1.0\linewidth]{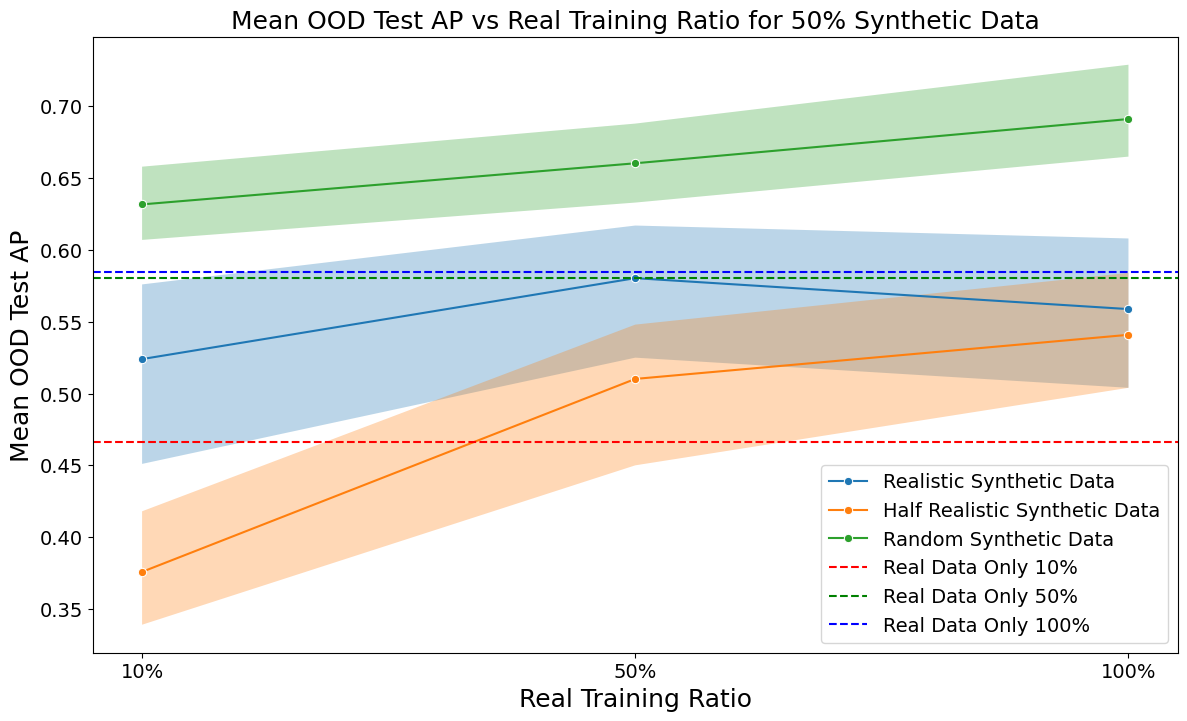}
        \caption{OOD Data Performance}
        \label{fig:mixedood}
    \end{subfigure}
    \caption{Faster R-CNN performance on the test and OOD dataset after fine-tuning the model on both real and \textbf{50\%} synthetic data per the \textit{mixed data} strategy.}
    \label{fig:mixeddata}
\end{figure}

While we cannot solely rely on synthetic data, the mixed data experiments indicate that adding synthetic data can \textit{reduce} the amount of real data needed.
\Cref{fig:mixeddata}, which displays the effect on a model's performance when injecting 50\% synthetic data into different ratios of real data, summarizes the key findings (see~\Cref{sec:appx_results} for all results).
Most notable is the performance boost when mixing any form of synthetic data with 10\% real data for both the in-distribution and OOD tests.
These positive effects are not limited to the low-resource scenario, as intertwining synthetic data with higher ratios of real-world data leads to better test and OOD performance when compared to the real-data-only counterpart.

A trade-off is also observed between test and OOD performance:
Half-realistic data yields strong performance on the test dataset but (at best) minimally impacts OOD performance, whereas random synthetic data shows the opposite effect—higher impact on OOD performance with minimal test performance impact.
Realistic synthetic data generally falls between these two extremes (cf.~\cite{Vanherle2022BMVC}).

\subsection{Bridged Transfer Learning}
\begin{figure}[h]
    \centering
    \begin{subfigure}[b]{0.49\linewidth}
        \centering
        \includegraphics[width=1.0\linewidth]{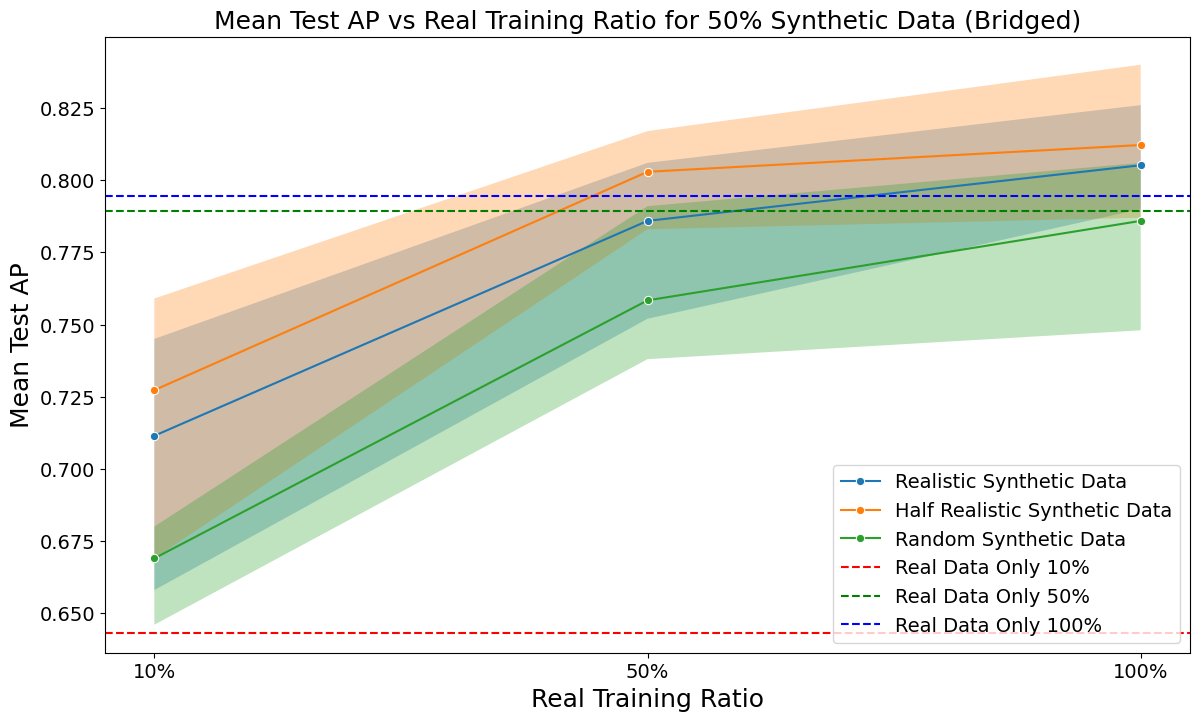}
        \caption{Test Performance}
        \label{fig:bridgedtest}
    \end{subfigure}
    \hfill
    \begin{subfigure}[b]{0.49\linewidth}
        \centering
        \includegraphics[width=1.0\linewidth]{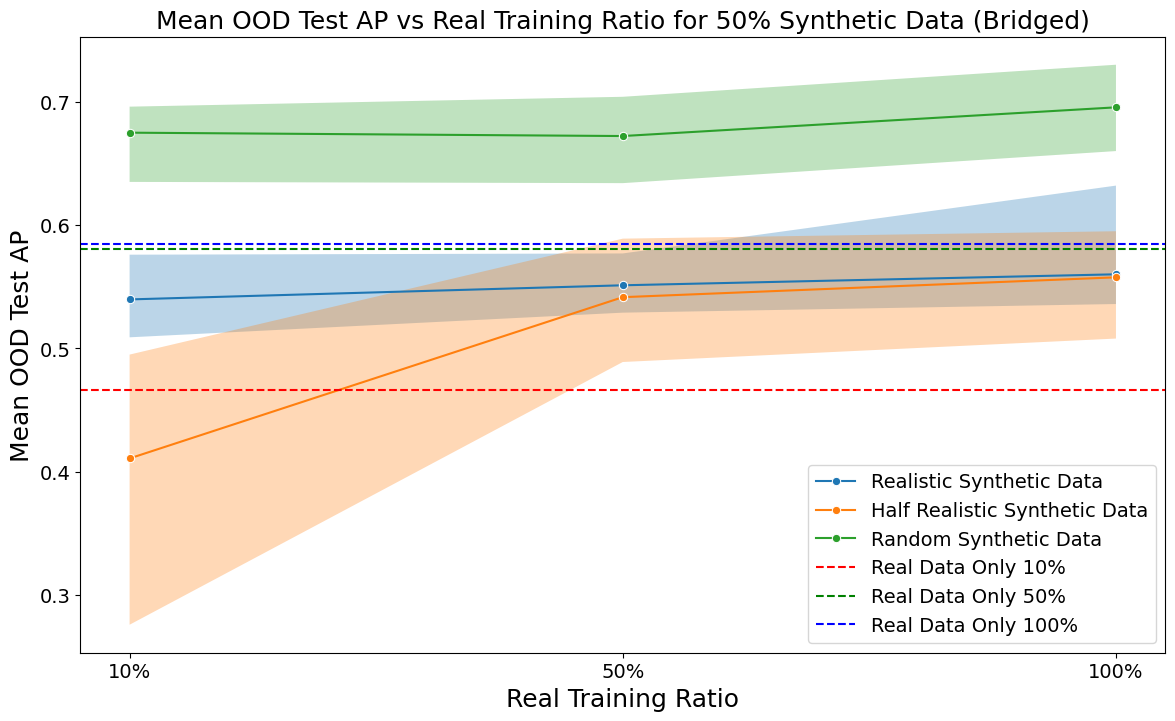}
        \caption{OOD Data Performance}
        \label{fig:bridgedood}
    \end{subfigure}
    \caption{Faster R-CNN performance on the test and OOD dataset after fine-tuning the model on both real and \textbf{50\%} synthetic data per \textit{bridged transfer learning}.}
    \label{fig:bridgeddata}
\end{figure}

As illustrated by~\cref{fig:bridgeddata} (see~\Cref{sec:appx_results} for all results), BTL generally preserves the core patterns observed for the mixed method.
Combining both data types is particularly beneficial when access to real data is limited.
BTL also exhibits the same trade-off between the different categories of synthetic data: half-realistic data improves test performance but weakens OOD capabilities, while random data improves OOD performance at the cost of poorer test performance.

Given fixed ratios for synthetic and real data and a fixed synthetic data type, Faster R-CNN models predominantly perform better when incorporating synthetic data via BTL.
In most cases, BTL-based fine-tuning leads to higher performance than the mixed method.
While these improvements are not as substantial as those reported in related work (e.g.,~\cite{li2024synthetic}), our results indicate that bridged transfer learning generally outperforms the mixed approach in our experiments, especially in low-real-data regimes.

\section{Discussions}

\paragraph{Synthetic-only training.}
The large performance gap on test data, despite more available synthetic images, suggests two non-exclusive explanations:
$(i)$ the synthetic data may be insufficiently information-rich for the target task;
$(ii)$ the synthetic distribution, even when realistic, may differ from our real distribution such that test data behaves like OOD relative to synthetic (and vice versa).
We also observe that randomly generated synthetic data produces the strongest OOD results among synthetic-only settings, consistent with domain randomization effects.

\paragraph{Trade-off between ID and OOD.}
Across settings, stronger test (ID) performance often coincides with weaker OOD performance and vice versa.
For example, half-realistic data produces the best test results and the weakest OOD results, whereas random data reverses this pattern.
We also note a drop in OOD performance when introducing large amounts of \textit{realistic} synthetic data, which contrasts with the generally improving OOD performance observed at higher real-data ratios.
These observations align with the idea that diversity and abstraction (as in random data) act as domain randomization, improving robustness at some cost to ID fit.

\paragraph{Mixed vs.\ BTL.}
Both mixed and BTL approaches are beneficial in low-real-data regimes.
BTL tends to yield higher performance than the mixed method under matched ratios and synthetic types, while preserving the same ID/OOD trade-offs across synthetic variants.
In practice, this suggests using BTL when feasible, while selecting the synthetic variant according to the deployment objective (ID accuracy vs. OOD robustness).

\paragraph{Limitations.} Our study uses a single class (pallets), a single detector (Faster R-CNN), and small real test sets (20 in-distribution and 20 OOD images). Effects are modest and sometimes within run-to-run variance across 10 seeds; we therefore treat differences as descriptive rather than definitive. Extending to more detectors, classes, and larger test sets is left for future work.
Furthermore, our OOD set was hand-curated from a single site.
Hence, its shift factors (color/orientation/occlusion) may not accurately represent all deployment conditions.

\section{Conclusion}
We studied synthetic data for fine-tuning a pre-trained Faster R-CNN on single-class pallet detection, comparing it with real-world data. 
In our small-scale setup, combining synthetic with limited real data yielded modest gains in in-distribution accuracy and out-of-distribution robustness, sometimes within run-to-run variability, while synthetic data alone was insufficient to replace real data; bridged transfer generally performed best. 
Across three variants – Realistic, Half-Realistic, and Random – we observed a trade-off: 
Half-Realistic tended to boost in-distribution performance, whereas Random improved OOD robustness, underscoring the value of domain randomization.
Overall, among the synthetic variants, Half-Realistic offered the strongest in-distribution gains (especially in low-data regimes), while Random strengthened OOD performance. 
For practitioners with limited real-world data, a well-designed synthetic pipeline can be both cost- and time-efficient, and once established, adaptable to new classes and conditions. 
These findings are descriptive rather than definitive; scaling to additional detectors, categories, and larger test sets, as well as integrating complementary generative approaches, remains a promising direction.

\bibliographystyle{splncs04}
\bibliography{egbib}
\newpage
\appendix

\section{Additional Experimental Results}
\label{sec:appx_results}

\begin{figure}[h]
    \centering
    \begin{subfigure}[b]{0.49\linewidth}
        \centering
        \includegraphics[width=1.0\linewidth]{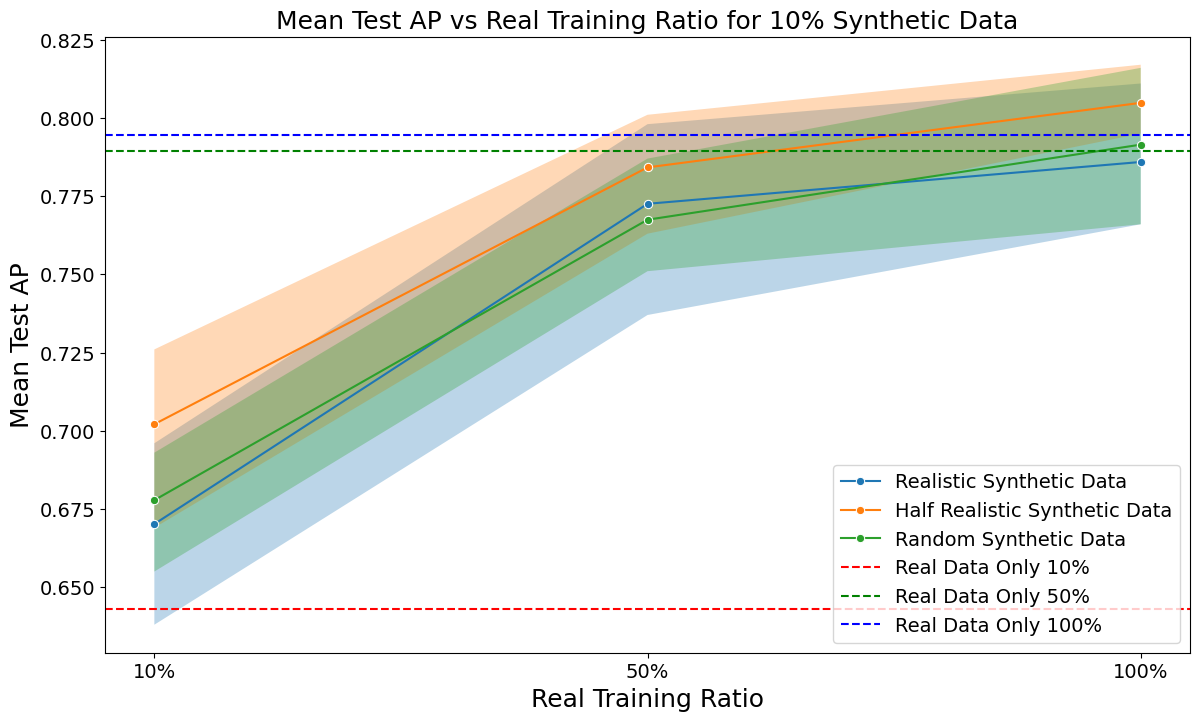}
        \caption{Test Performance (10\%)}
        \label{fig:mixed10test}
    \end{subfigure}
    \hfill
    \begin{subfigure}[b]{0.49\linewidth}
        \centering
        \includegraphics[width=1.0\linewidth]{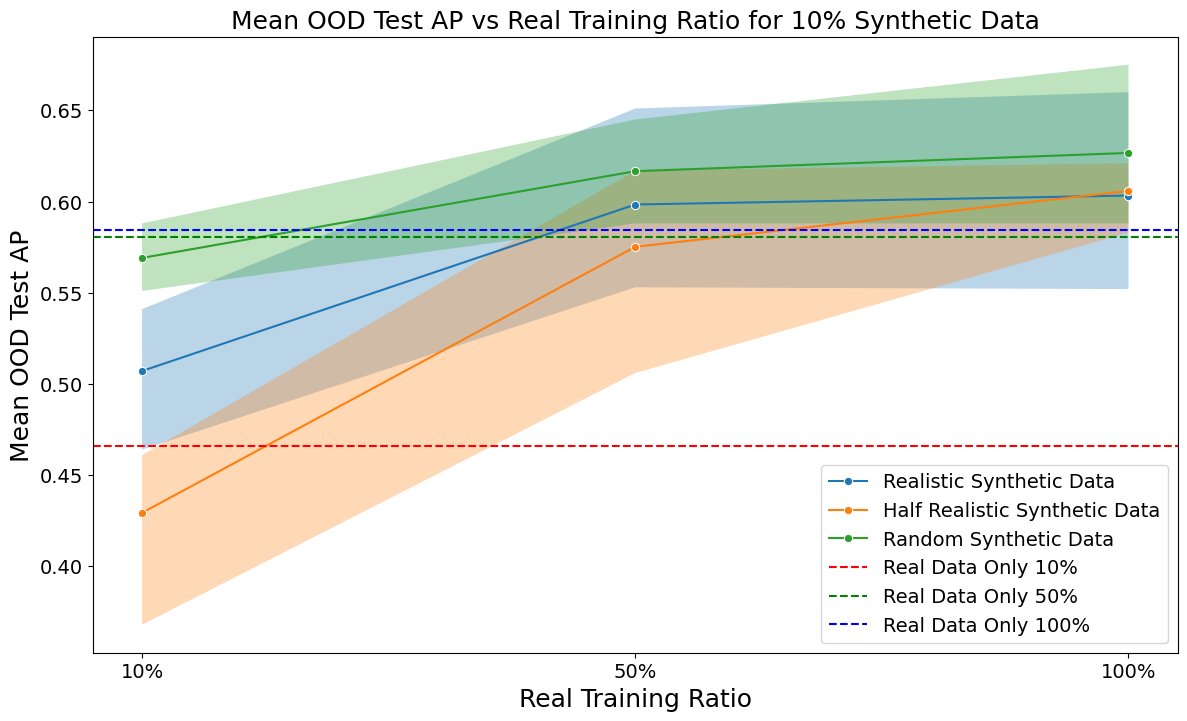}
        \caption{OOD Data Performance (10\%)}
        \label{fig:mixed10ood}
    \end{subfigure}
    \begin{subfigure}[b]{0.49\linewidth}
        \centering
        \includegraphics[width=1.0\linewidth]{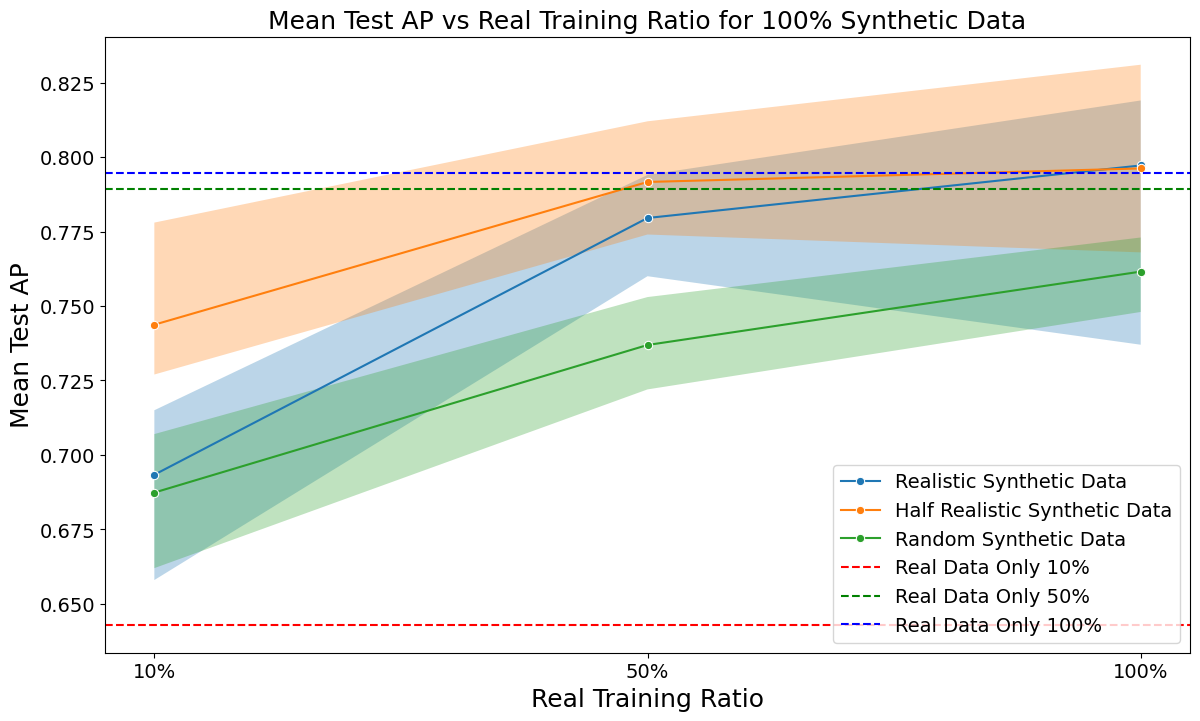}
        \caption{Test Performance (100\%)}
        \label{fig:mixed100test}
    \end{subfigure}
    \hfill
    \begin{subfigure}[b]{0.49\linewidth}
        \centering
        \includegraphics[width=1.0\linewidth]{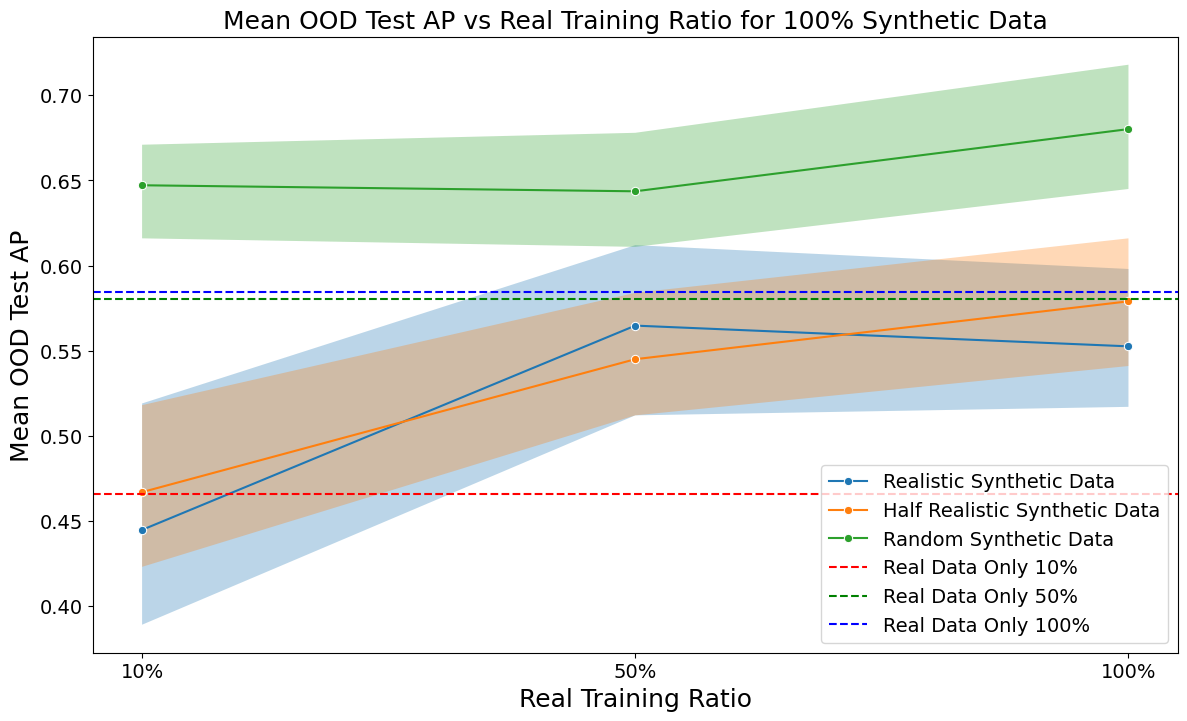}
        \caption{OOD Data Performance (100\%)}
        \label{fig:mixed100ood}
    \end{subfigure}
    \caption{Faster R-CNN performance on the test and OOD dataset after fine-tuning the model on both real and \textbf{10\%} (\cref{fig:mixed10test,fig:mixed10ood}) resp. \textbf{100\%} (\cref{fig:mixed100test,fig:mixed100ood}) synthetic data per the \textit{mixed data} strategy.}
    \label{fig:mixedrest}
\end{figure}

\begin{figure}[h]
    \centering
    \begin{subfigure}[b]{0.49\linewidth}
        \centering
        \includegraphics[width=1.0\linewidth]{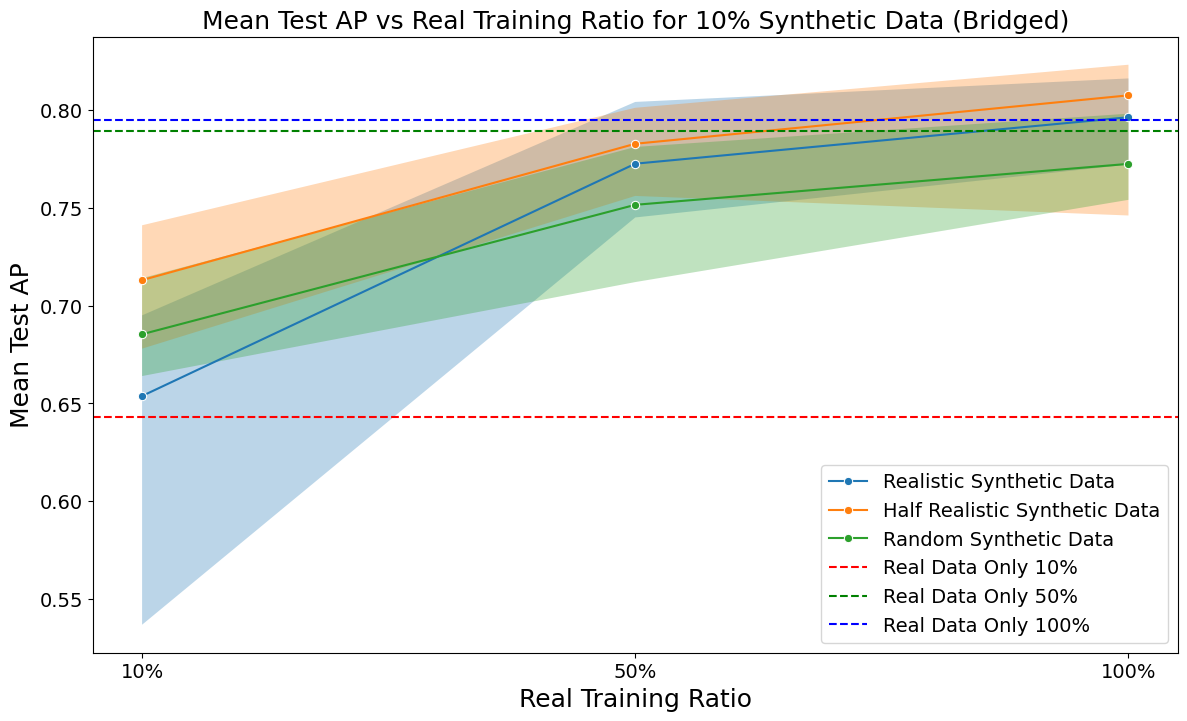}
        \caption{Test Performance (10\%)}
        \label{fig:bridged10test}
    \end{subfigure}
    \hfill
    \begin{subfigure}[b]{0.49\linewidth}
        \centering
        \includegraphics[width=1.0\linewidth]{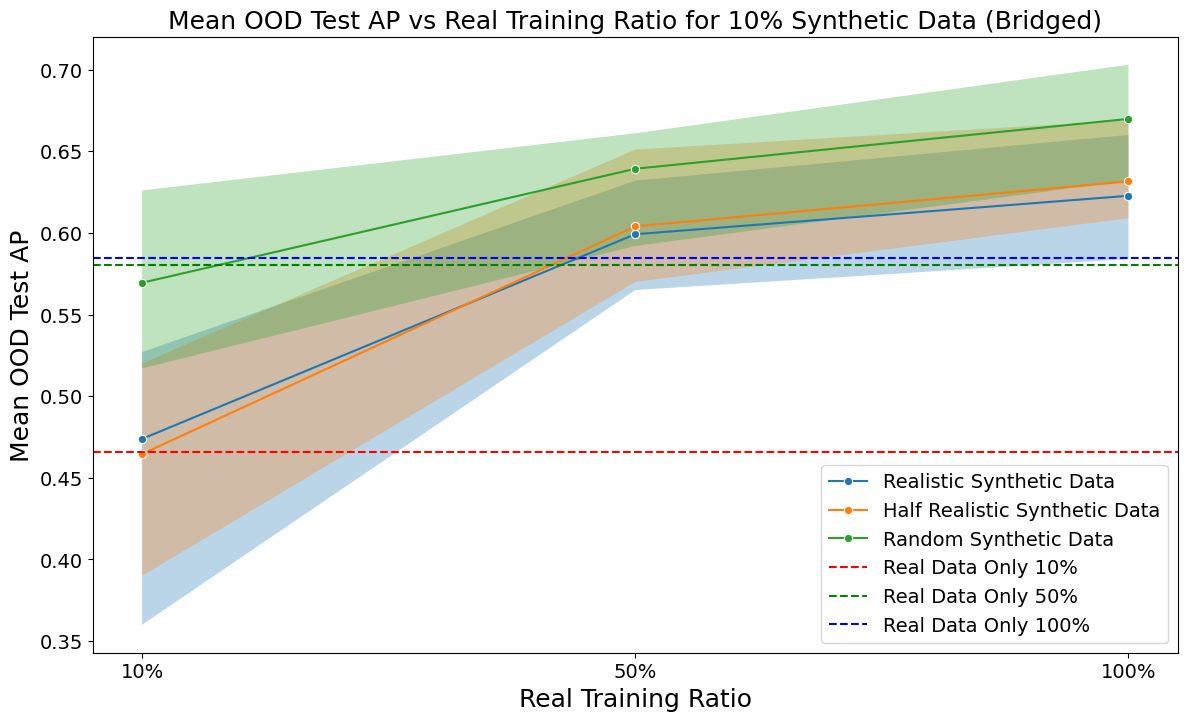}
        \caption{OOD Data Performance (10\%)}
        \label{fig:bridged10ood}
    \end{subfigure}
    \begin{subfigure}[b]{0.49\linewidth}
        \centering
        \includegraphics[width=1.0\linewidth]{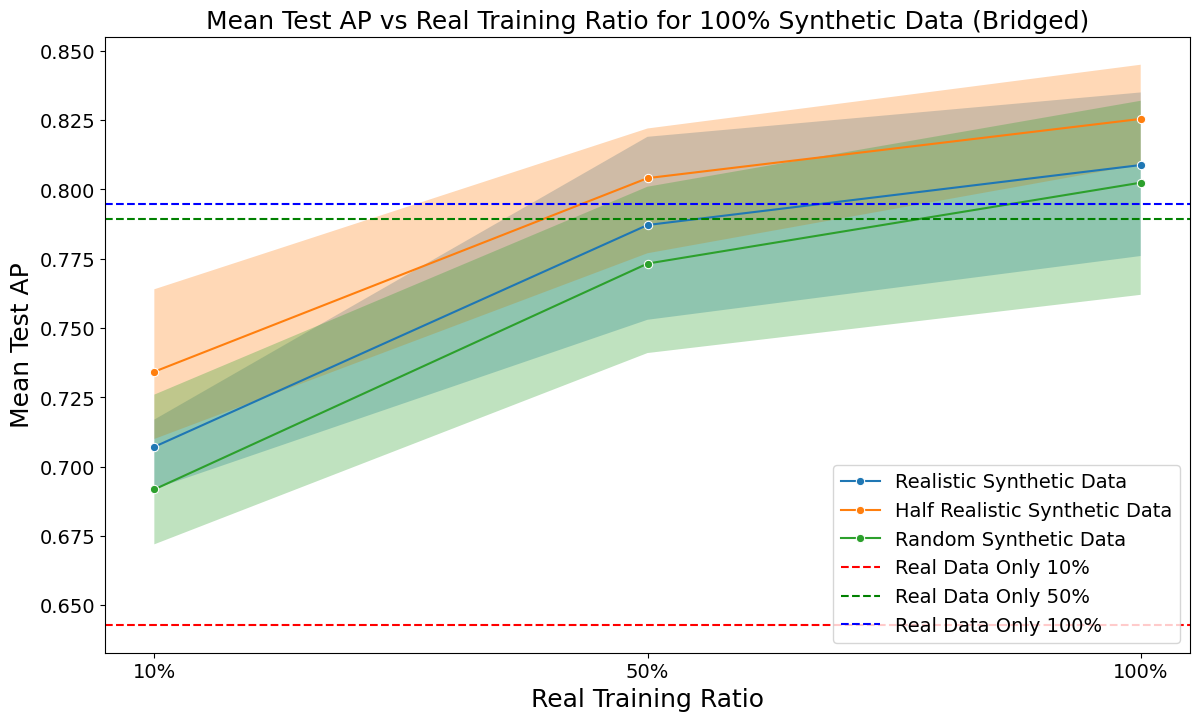}
        \caption{Test Performance (100\%)}
        \label{fig:bridged100test}
    \end{subfigure}
    \hfill
    \begin{subfigure}[b]{0.49\linewidth}
        \centering
        \includegraphics[width=1.0\linewidth]{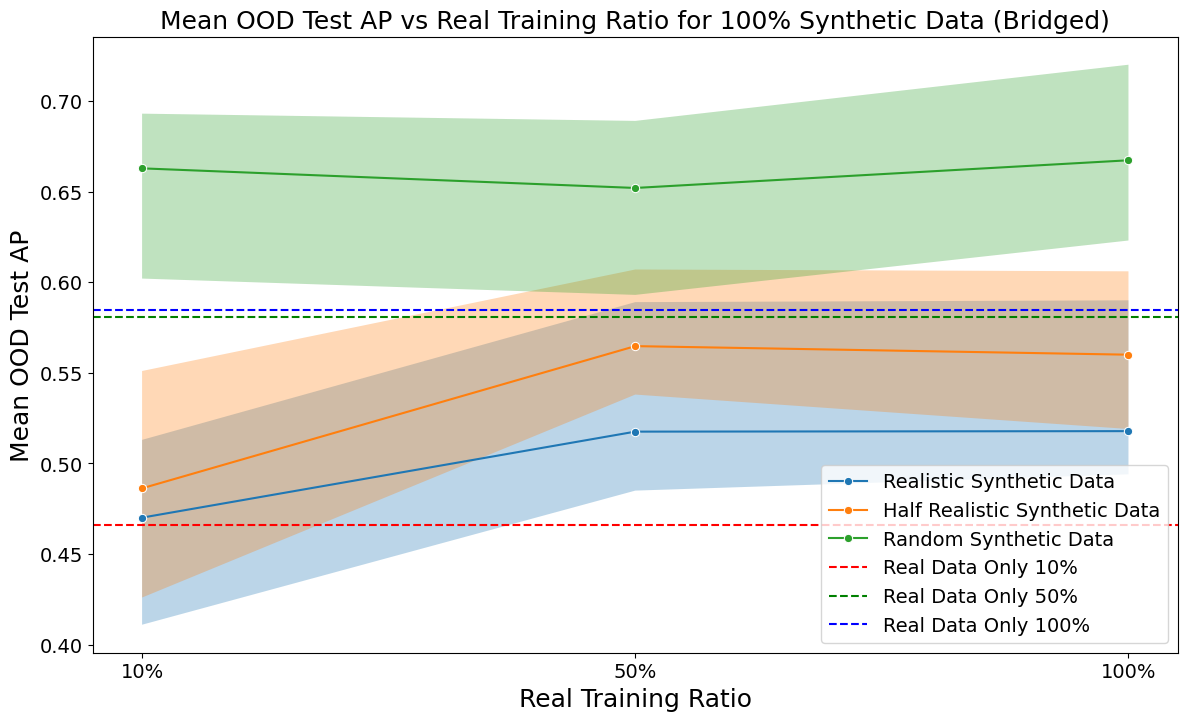}
        \caption{OOD Data Performance (100\%)}
        \label{fig:bridged100ood}
    \end{subfigure}
    \caption{Faster R-CNN performance on the test and OOD dataset after fine-tuning the model on both real and \textbf{10\%} (\cref{fig:bridged10test,fig:bridged10ood}) resp. \textbf{100\%} (\cref{fig:bridged100test,fig:bridged100ood}) synthetic data per \textit{bridged transfer learning}.}
    \label{fig:bridgedrest}
\end{figure}

\end{document}